# Explainability Fact Sheets:
# A Framework for Systematic Assessment of Explainable Approaches

Kacper Sokol
K.Sokol@bristol.ac.uk
Department of Computer Science, University of Bristol
Bristol, United Kingdom

Peter Flach
Peter.Flach@bristol.ac.uk
Department of Computer Science, University of Bristol
Bristol, United Kingdom

## ABSTRACT

Explanations in Machine Learning come in many forms, but a consensus regarding their desired properties is yet to emerge. In this paper we introduce a taxonomy and a set of descriptors that can be used to characterise and systematically assess explainable systems along five key dimensions: functional, operational, usability, safety and validation. In order to design a comprehensive and representative taxonomy and associated descriptors we surveyed the eXplainable Artificial Intelligence literature, extracting the criteria and desiderata that other authors have proposed or implicitly used in their research. The survey includes papers introducing new explainability algorithms to see what criteria are used to guide their development and how these algorithms are evaluated, as well as papers proposing such criteria from both computer science and social science perspectives. This novel framework allows to systematically compare and contrast explainability approaches, not just to better understand their capabilities but also to identify discrepancies between their theoretical qualities and properties of their implementations. We developed an operationalisation of the framework in the form of *Explainability Fact Sheets*, which enable researchers and practitioners alike to quickly grasp capabilities and limitations of a particular explainable method. When used as a *Work Sheet*, our taxonomy can guide the development of new explainability approaches by aiding in their critical evaluation along the five proposed dimensions.

## 1 INTRODUCTION

With the current surge in eXplainable Artificial Intelligence (XAI) research it has become a challenge to keep track of, analyse and compare many of these approaches. A lack of clearly defined properties that explainable systems should be evaluated against hinders progress of this fast moving research field because of undiscovered (or undisclosed) limitations and properties of explainability approaches and their implementations (as well as discrepancies between the two).

In this work we propose that every explainability method designed for predictive systems should be accompanied by a *Fact Sheet* that assesses its (1) *functional* and (2) *operational* requirements. Moreover, the quality of explanations should be evaluated against a list of (3) *usability* criteria to better understand their usefulness from a user perspective. Their (4) *security*, *privacy* and any *vulnerabilities* that they may introduced to the predictive system should also be discussed on this *Fact Sheet*. Lastly, their (5) *validation*, either via user studies or synthetic experiments, should be disclosed. Therefore, a standardised list of explainability properties – or, to put it otherwise, desiderata – spanning all of these five dimensions would facilitate a common ground for easy evaluation and comparison of explainability approaches, and help their designers consider a range of clearly defined aspects important for this type of techniques.

Despite theoretical guarantees for selected explainability approaches, some of these properties can be lost in implementation, due to a particular application domain or data set used. As it stands, many implementations do not exploit the full potential of the selected explainability technique, for example, a method based on counterfactuals might not take advantage [53] of their social and interactive aspects [34]. A use of guidelines or a systematic evaluation of an approach with a standardised *Fact Sheet* could help discover these unexpected functionality losses and account for or, simply, report them for the benefit of the research community and end users.

To mitigate the current lack of consensus regarding a common set of properties that every explainable method should be evaluated against, in this paper we collect, review and organise a comprehensive set of characteristics that span both computer science and social sciences insights on that matter. Our goal is to provide the community with an *Explainability Fact Sheet* template that users of explainable systems (researchers and practitioners) could utilise to systematically discuss, evaluate and report properties of their techniques. This will not only benefit research, but will be of particular importance as a guideline for designing, deploying or evaluating explainability methods, especially if compliance with best practices or legal regulations is required, e.g., the "right to explanation" introduced by the European Union's General Data Protection Regulation (GDPR) [53]. Therefore, creators of explainable methods who need to understand their specific requirements can use the same template as a *Work Sheet*. We demonstrate the usefulness and practical aspects of these *Explainability Fact Sheets* by presenting a particular instantiation created for the Local Interpretable Model-agnostic Explanations (LIME) algorithm [44] (see the supplementary material).

## 2 EXPLAINABILITY FACT SHEETS DIMENSIONS

*Explainability Fact Sheets* are intended to evaluate explanatory systems along five dimensions. Firstly, the *functional* one, which considers algorithmic requirements such as the type of problem it applies to (classification, regression, etc.), which component of the predictive system it is designed for (data, models or predictions), the scope of the technique (e.g., local vs. global explanation) and its relation to the predictive system (post-hoc vs. ante-hoc), among many others. Secondly, the *operational* one, which includes the type of interaction with the end user, explainability vs. predictive performance trade-off and the user's background knowledge required to fully utilise the explainability power of a technique, to name a few. Thirdly, the *usability* one, which takes a user-centred perspective and includes properties of the explanation that make it feel "natural" and easily comprehensible by the explainee. To fulfil the user's expectations these include: fidelity of the explanation, its actionability from the explainee's perspective and its brevity, to give just a few examples. The fourth one is *safety*, which discusses robustness and security of an explainability approach. For example, these include an analysis of how much information about the predictive model or its training data an explanation leaks and whether an explanation for a fixed data point is consistent throughout different models given the same training data. Finally, the fact sheet discusses the *validation* process used to evaluate and prove the effectiveness of an explainability approach by describing a user study or synthetic approach that was carried out. Since the *Fact Sheet* can evolve over time, we encourage their creators to systematically *version* them [2], thereby making their recipients aware of any updates. Furthermore, indicating whether the whole *Fact Sheet* or some of its parts are with respect to a (theoretical) algorithmic approach, an actual implementation or a mixture of the two will benefit its clarity. To put these explainability dimensions into context, an exemplar of the *Explainability Fact Sheet* for LIME [43] is provided as a supplementary material.

### 2.1 Functional Requirements

These can help to determine *whether a particular approach is suitable for a desired application*, therefore they resemble a taxonomy of Machine Learning (ML) and explainability approaches. The list provided below can be thought of as a check-list of an engineer who was tasked with identifying and deploying the most suitable explainability algorithm for a particular use case. All of these properties are well-defined and flexible enough to accommodate any explainability approach.

**F1** *Problem Supervision Level.* An explainability approach can be applicable to any of the following learning tasks: *unsupervised*, *semi-supervised*, *supervised* and *reinforcement*. A large part of the literature focuses on supervised learning, where explanations serve as a justification of a prediction. Nevertheless, explainability can also benefit unsupervised learning, where the user may want to learn about the data insights elicited by a model; reinforcement learning, where the user is interested in autonomous agent's decisions; and semi-supervised learning, where the user can help the system choose the most informative data points for learning by understanding the system's behaviour.

**F2** *Problem Type.* We can identify three main problem types in Machine Learning: *classification* (binary/multi-class/multi-label and probabilistic/non-probabilistic), *regression*, and *clustering*. (Additionally, we can consider types such as ranking and collaborative filtering.) By clearly defining the applicable type of a learning task for an explainability method, potential users can easily identify ones that are useful to them.

**F3** *Explanation Target.* The Machine Learning process has three main components: *data* (both raw data and features), *models* and *predictions*. Explaining the first one may be difficult or even impossible without any modelling assumptions. These are usually summary statistics, class ratio, feature plots, feature correlation and dimensionality reduction techniques. Explaining models is concerned with its general functionality and conveying its conceptual behaviour to the explainee. Explaining predictions provide a rationale behind the output for any particular data point.

**F4** *Explanation Breadth/Scope.* This notion varies across data, models and predictions. It tells the user to what extent an explanation can be generalised. (See **U3** for a complementary view on this property from the *usability* perspective.) The main three explanation generalisability stages are: *local* – a single data point or a prediction, *cohort* – a subgroup in a data set or a subspace in the model's decision space, and *global* – a comprehensive model explanation.

**F5** *Computational Complexity.* Given that some applications may have either *time*, *memory* or *computational power* constrains, each explainability approach should consider these. If, for example, a given method is capable of explaining both a single prediction and the whole model, both of these aspects should be discussed. *Algorithmic complexity* measures such as Big-O or Little-O notations can be used to assess these aspects of an explainable system. Alternatively to theoretical performance bounds, empirical evaluation can also be discussed, e.g., the average time over 1,000 iterations that it took to generate an explanation for a single data point with fixed parameters of the explainability algorithm on a single-core CPU with 16GB of RAM.

**F6** *Applicable Model Class.* We can identify three main degrees of *portability* [45] for explainability algorithms: *model-agnostic* – working with any model family; *model class-specific* – designed for a particular model family, e.g., logical or linear models; and *model-specific* – only applicable to a particular model, e.g., decision trees.

**F7  Relation to the Predictive System.** We can characterise two main relations between a predictive model and an explainability technique. *Ante-hoc* approaches use the same model for predicting and explaining, e.g., explaining a linear regression with its feature weights. It is important to note that some of these techniques may come with caveats and assumptions about the training data or the training process (**F9**), which need to be satisfied for the explanation to work as intended. On the other hand, in *post-hoc* approaches predictions and explanations are made with different models, e.g., a local surrogate explainer. One can also name a third type – a special case of the post-hoc family – a (*global*) *mimic approach*, where to explain a complex (black-box) model a simpler (inherently transparent) model is built in an attempt to mimic the behaviour of the more complex one, e.g., a global surrogate explainer.

**F8  Compatible Feature Types.** Not all models are capable of handling all the feature types, for example, categorical features are at odds with predictive algorithms that use optimisation as their backbone. Furthermore, selected model implementations require categorical features to be pre-processed, one-hot encoded for example, rendering them incompatible with some explainability approaches. Therefore, every method should have a clear description of compatible feature types: *numerical*, *ordinal* (we differentiate numerical and ordinal features as the latter may have a bounded range) and *categorical*. In addition to these standard feature types, some tasks may come with a hierarchy of features and/or their values, in which case the explainability algorithm should clearly state whether these are beneficial for the quality of the explanation and how to utilise this information.

**F9  Caveats and Assumptions.** Finally, any *functional* aspects of an explainability approach that do not fit into these categories should be included at the end. In particular, restrictions with respect to input and output of predictive models and explainability techniques [36], e.g., support for black-and-white images only, validated behaviour on text corpora up to 100 tokens, numerical confidence of a prediction or an explanation, assumptions such as *feature independence*, the effect of correlated features on the quality of an explanation, or, simply, explainability technique-specific requirements such as *feature normalisation* when explaining a linear model with its weights.

## 2.2  Operational Requirements

The following properties characterise *how users interact with an explainable system and what is expected of them*. They can be thought of as considerations from a deployment point of view.

**O1  Explanation Family.** A great categorisation, which we believe is still up to date, of explainable approaches accounting for their presence in philosophy, psychology and cognitive science was introduced by Johnson and Johnson [18] for expert systems. The authors have identified three main types of explanations: *associations between antecedent and consequent*, e.g., model internals such as its parameters, feature(s)-prediction relations such as explanations based on feature attribution or importance and item(s)-prediction relations [24] such as influential training instances [23] (or neighbouring data points); *contrasts and differences* (using examples), e.g., prototypes and criticisms [20, 21] (similarities and dissimilarities) and class-contrastive counterfactual statements [34]; and *causal mechanisms*, e.g., a full causal model [38].

**O2  Explanatory Medium.** An explanation can be delivered as a: *(statistical) summarisation*, *visualisation*, *textualisation*, *formal argumentation* or mixture of the above. Examples of the first one are usually given as numbers, for example, coefficients of a linear model or a summary statistics of a data set. The second one comprises all sort of plots that can be used to help the user comprehend a predictive system's behaviour, e.g., Individual Conditional Expectation [12] or Partial Dependence [10] plots. The textualisation is understood as any explanation in form of a natural language description, e.g., a dialogue system that an explainee can query. An explanation system based on a formal argumentation framework [8] encompasses approaches that can output logical reasoning behind an explanation, hence provide the explainee with an opportunity to argue against it, whether in a form of a natural language conversation or highlighting important regions in an image. Finally, an example of a mixture of these representations can be a plot accompanied by a caption that helps to convey the explanation to the end user. Such a mixture may be necessary at times as not all of the media are able to convey the same amount or type of information [9]. For example, visualisations are confined to three dimensions (four when counting time, i.e., animations) due to the limitations of the human visual perception system and counterintuitiveness of higher dimensions – a phenomenon known as the curse of dimensionality. The choice of an explanatory medium is also important as it may limit the *expressive power* of an explanation.

**O3  System Interaction.** The communication protocol that an explainable method employs can be either *static* or *interactive* (with and without user feedback). The first one is a one-size-fits-most approach where the system outputs an explanation based on a pre-defined protocol specified by the system designer, hence may not always satisfy the user's expectations, for example, always outputting the most significant factors in favour and against a particular prediction when the user is interested in a feature not listed therein. Alternatively, the system can be interactive, thereby allowing an explainee to explore all aspects of a particular explanation. These include interactive web interfaces and dialogue systems, amongst others. Furthermore, in case of an interactive system, its creator should indicate whether the explainer can incorporate any feedback (and in what form) given by the explainee and how, if at all, it influences the underlying predictive model (e.g., incremental-learning algorithms) [26].

**O4  Explanation Domain.** Explanations are usually expressed in terms of the underlying model's parameters or data exemplars and their features – *original domain*. However, it may be the case that the explanation is presented in a different form – *transformed domain*. Consider, for example, a system explaining image classification results where the data domain is an image and the explanation is a natural language description as opposed to a saliency map overlaid on top of the image. Another approach can be an *interpretable data representation*, e.g., super-pixels instead of raw pixel values, introduced by Ribeiro et al. [44] as part of the LIME algorithm (cf. **O5**).

**O5**  *Data and Model Transparency.*  An explainable approach should clearly indicate whether the underlying predictive algorithm and/or the (training) data should be *transparent* or they can be *opaque*. (This requirement is tightly related to **F7**, in particular when we are dealing with ante-hoc explainability approaches.) In case of model explanations, does an explainee need to understand the inner workings of a predictive model? When data or predictions are being explained, do the data features need to be human understandable in the first place? (This concept, in turn, is related to Lipton's [32] validation approaches discussed in the last paragraph of Section 2.5: what sort of understanding of the model and/or features is expected of the user.) For example, consider explaining a prediction in terms of a room temperature vs. using a squared sum of a room temperature and its height. In cases where the input domain is not human comprehensible, the system designer may want to give a list of data transformations as a remedy or choose an example-based explanation [48]. For example, applying a super-pixel segmentation to an image and using its output as higher-level features that are human-comprehensible can help to explain image classification tasks, which is done by LIME.

**O6**  *Explanation Audience.*  The intended audience of an explainable method may vary from a *domain expert*, through a requirement of a *general knowledge about a problem*, all the way to a *lay audience*. Considering the type of the domain expertise is also important: ML and AI knowledge vs. domain knowledge. Therefore, discussing the level and type of background knowledge required to comprehend an explanation is crucial [4, 41, 52]. Some techniques may allow to adjust (**U4** and **U5**) the size (**U11**) or the complexity level (**U9**) of an explanation based on the audience (**U10**) via some of the *usability requirements* (Section 2.3). Furthermore, the transparency of the features (and/or the model, cf. **O5**) should be considered with respect to the intended audience. For example, consider a system that explains its predictions using natural language sentences. Given the language skills of the recipient, the system can use sentences of varying grammatical and vocabulary complexity to facilitate their easier comprehension [54]. Finally, given the cognitive capacity of an explainee, the system may be able to adjust the granularity of an explanation to suit the recipient's needs. For example, explaining a disease diagnosis to doctors as opposed to patients or their families. One of the goals of decreasing the complexity of an explanation (which is sometimes at odds with its fidelity, cf. **U1** and **U2**) may be making it easy enough for the explainees to comprehend it in full [32], hence enable them to simulate the decisive process *in vivo*, i.e., simultability. (See the validation requirements discussed in the last paragraph of Section 2.5 for a detailed description of this concept.)

**O7**  *Function of the Explanation.*  Every explainability approach should be accompanied by a list of its intended applications [24]. Most of them are designed for transparency: *explaining* a component of the ML pipeline to an end user (whether it is to support decisions, compare model, elicit knowledge form a predictive model or the data used to build it, or extract a causal relation). Nevertheless, some of them can also be used to assess *accountability* of the underlying predictive model, e.g., debug and diagnose it to engender trust; or demonstrate its *fairness* (e.g., disparate treatment via counterfactuals). It is important to provide the user with the envisaged (and validated) deployment context to prevent its misuse, which may lead to an unintentional harm when deployed in high-risk applications or autonomous system.

**O8**  *Causality vs. Actionability.*  Most explanations are not of a causal nature. If this is the case, this property needs to be explicitly communicated to the users so that they can avoid drawing incorrect conclusions. Given an *actionable*, and not causal, explanation, the users should understand that the action suggested by the explanation will result in, say, a different classification outcome, however interpreting it causally, no matter how tempting, can lead to inaccurate insights [38]. Similarly, explanations that are derived from a full causal model should be advertised as *causal* and used to their full potential. This concept is closely related to **O1**, which specifies the explanation family.

**O9**  *Trust vs. Performance.*  All of the explainability approaches should be accompanied by a critical discussion of performance–explainability trade-offs that the user has to face. By large, explainability can improve users' *trust* in a predictive system, nevertheless sometimes the decrease in *predictive performance* that is associated with making a particular system more explainable may not be worth it. For example, consider a case where making a predictive algorithm explainable renders its predictions to be incorrect most of the time. Therefore, the user of an explainability method needs to decide whether the main objective of a predictive system is to make it more efficient or learn something from data.

**O10**  *Provenance.*  Finally, an *Explainability Fact Sheet* should be translucent [45] about the information that contribute to every explanation outputted by an explainability method. Most often provenance of an explanation can be attributed to its reliance on: a *predictive model* – achieved via interacting with the (black-box) model or using its internal parameters (glass-box) [24]; a *data set* – introduced by inspecting or comparing data points originating from one or a mixture of the training, evaluation and validation data sets; or, ideally, *both* a predictive model and a data set. An example of a purely model-driven explanation is interpreting a $k$-means model with its centroids. A purely data-driven explanation is, for example, explaining predictions of a $k$-nearest neighbours model by accounting only for the $k$ closest neighbours to the data point being explained. If possible, every explanation should be accompanied by an *explainability trace* indicating which training data points were influential for a prediction [23] and the role that the model and its parameters played. In most of the cases, a model-specific explainability algorithm (**F6**) will heavily rely on internal parameters of the underlying predictive model, whereas a model-agnostic approach will rely more on data and behaviour of a predictive model.

## 2.3  Usability Requirements

Here, we discuss *properties of explanations that are important from an explainee's point of view*. Many of these are grounded in social science research, hence, whenever applicable, making algorithmic explanations feel more natural to the end users regardless of their background knowledge and prior experience with this type of systems or technology in general.

**U1   Soundness.** This property measures how *truthful* an explanation is with respect to the underlying predictive model [26] (sometimes called concordance). Its goal is to quantify local or global adherence of the explanation to the underlying model. If the explanation is of *ante-hoc* type, this property does not apply as both the explanation and the prediction are derived using the same algorithm. However, explanations based on *post-hoc* (or *mimic*) approaches should measure this property to quantify the error introduced by the explainability technique. This can be done by calculating a selected performance metric between the outputs of predictive and explanatory models, for example, average rank correlation between the two. A high value of such a metric would ensure the user that an explanation is consistent and aligned with predictions of the underlying model. This requirement can also be understood as "truthfulness" of an explanation – Grice et al. [13] have noted in their *maxim of quality*, which is one of the rules for cooperative communication, that a user should only be presented with claims supported by evidence. Soundness is usually one of the two explanation *fidelity* measures, with the other one being *completeness* (**U2**).

**U2   Completeness.** For an explanation to be trusted, it also needs to *generalise* well beyond the particular case in which it was produced. This mostly applies to *local* and *cohort* explanations as the user may want to apply insights learnt from one of these explanations to a similar case: *pars pro toto*. Completeness measures how well does an explanation generalise [28, 33, 35], hence to what extent it *covers* the underlying predictive model. It can be quantified by checking correctness of an explanation across similar data points (individuals) across multiple groups within a data set. In particular, a support set – the number of instances to which the explanation applies divided by the total number of instances – can be used to measure completeness. Given a context-dependent nature of this metric, there is no silver bullet to assess how well an explanation *encompasses* the model. In addition to *soundness*, this is the second explanation *fidelity* metric.

**U3   Contextfullness.** If there are known issues with an explanation completeness, a user may not trust it. To overcome this and help the user better understand how an explanation can be generalised, it can be framed in a *context*, thereby allowing the user to assess its *soundness* and *completeness*. For example, the user will better understand the limitations of an explanation if it is accompanied by all the necessary conditions for it to hold, critiques (i.e., explanation oddities) and its similarities to other cases [20, 21, 26, 34]. Contextfullness can help to make a local explanation either explicitly local, allow the explainee to safely generalise it to a cohort-based explanation, or even indicate that despite it being derived for a single prediction it can be treated as a global one. A specific (quantitative) case of this property, called *representativeness*, aims to measure how many instances in a (validation) data set does a single explanation cover. Another aspect of contextfullness is the degree of importance for each factor contributing to an explanation. For example, if an explanation is supported by three causes, how important are they individually. One observation worth noting is that an order in which they are presented rarely ever indicates their relative importance, e.g., a list of conditions in a decision rule. This usability requirement can also be compared to the *maxim of manner* (from the rules for cooperative communication [13]) that entails being as clear as possible in order to avoid any ambiguity that may lead to a confusion at any point.

**U4   Interactiveness.** Given a wide range of explainees' experience and background knowledge, one explanation cannot satisfy their wide range of expectations. Therefore, to improve the overall user experience the explanation process should be controllable. For example, it should be *reversible* (in case the user inputs a wrong answer), respect *user's preferences and feedback*, be "*social*" (bidirectional communication is preferred to one-way information offloading), allow to adjust the *granularity* of an explanation and be *interactive* [20, 21, 26, 28, 34, 52, 55]. This means that, whenever possible, the users should be able to *customise* the explanation that they get to suit their needs [48]. For example, if the system explains its decisions with counterfactual statements and a foil used in such a statement does not contain information that the users are interested in, they should be able to request an explanation that contains this particular foil (if one exists).

**U5   Actionability.** When an explanation is provided to help users understand a reason behind a decision, then the users prefer explanations that they can treat as *guidelines* towards the desired outcome [25, 51]. For example, in a banking context, given an explanation based on counterfactual statements, it is better (from the user's perspective) to get a statement conditioned on a number of active loans rather than the user's age. The first one provides the user with an action towards the desired outcome (i.e., pay back one of the loans before reapplying), while the latter leaves the user without any options.

**U6   Chronology.** Some aspects of an explanation may have inherent time ordering, for example, loans taken by a borrower. In such cases, if one of the reasons given in an explanation has a time-line associated with it, users prefer explanations that account for *more recent* events as their cause, i.e., proximal causes [34]. For example, consider multiple events of the same type contributing equally to a decision: an applicant has three current loans and will not be given a new one unless one of the outstanding ones is paid back. Paying back any of the three loans is a sufficient explanation, however from the user's perspective taking the most recent loan is a more natural reason behind the unfavourable loan application outcome then having any of the first two loans.

**U7   Coherence.** Many users of explainable systems can have a prior background knowledge and beliefs about the matter that is being predicted and explained – their mental model of the domain. Therefore, any explainable system should be *consistent* with the explainee's prior knowledge [9, 33, 35], which can only be achieved when the explainee's mental model is part of the explainability system (**U10**) as otherwise there is nothing to be coherent with. (A mental model can either be *functional*, i.e., shallow, in which case the end users know how to interact with something but not how it works in detail, or *structural*, i.e., deep, in which case they have a detailed understanding of how and why something works [27].) If the prediction of a model is consistent with the users' expectations, then the reasoning behind it will not matter most of the time unless its logic is fundamentally flawed (internal inconsistency) or it is at odds with the general knowledge [54] – humans tend to ignore information that is inconsistent with their beliefs (confirmation bias). On the other hand, if the output of a predictive model is not what the users expect, they will contrast the explanation against their mental model to understand the prediction, in which case the explainer should identify and fill in these knowledge gaps [18]. Therefore, if an explanation uses arguments that are consistent with the users' beliefs, they will be more likely to accept it. This property is highly subjective, however a basic coherence with the universal laws, for example, number ordering, should be satisfied.

**U8** *Novelty.* Providing users with a mundane or expected explanation should be avoided. Explanations should contain surprising or abnormal characteristics (that have low probability of happening, e.g., a rare feature value) to point the user's attention in an interesting direction [3, 28, 34] (recall **U7** where anomalies prompt the user to request an explanation). However, this objective requires balancing the trade-off between coherence with the explainee's mental model, novelty and overall plausibility [54]. For example, consider an ML system where explanations help to better understand a given phenomenon. In this scenario, providing the users with explanations that highlight relations that they are already aware of should be avoided. The explainees' background knowledge should be considered before producing an explanation to make sure that it is novel and surprising; at the same time, consistency with their mental model should be preserved as long as it is correct. Again, this usability criterion can only be built on top of the explainees' mental model, since this knowledge is essential for assessing novelty of causes.

**U9** *Complexity.* Given a wide spectrum of skills and background knowledge of explainees, the *complexity* of explanations should be tuned to the recipient [34, 52]. This can be an operational property of an explainable system (**O6**), however it is also important to consider it from a user perspective. If the system does not allow for explanation complexity to be *adjusted* by the user, it should be as *simple* as possible by default (unless the explainee explicitly asks for a more complex one). For example, given an explanation of an automated medical diagnosis, it should use observable symptoms rather than the underlying biological processes responsible for the condition. Choosing the right complexity automatically may only be possible given the availability of the explainee's mental model.

**U10** *Personalisation.* Adjusting an explanation to users requires the explainability technique to model their background knowledge and mental model [48, 55]. This is particularly important when attempting to adjust complexity of an explanation (**U9**), its novelty (**U8**) and coherence (**U7**). An explanation can either be personalised on-line via an interaction or off-line by incorporating the necessary information into the model (e.g., parameterisation) or data. Personalising an explanation is related to another rule of cooperative communication [13]: the *maxim of relation*. According to this rule, a communication should only relay information that are relevant and necessary at any given point in time. Therefore, an explainability system has to "know" what the user knows and expects to determine the content of the explanation [9].

**U11** *Parsimony.* Finally, explanations should be *selective* and *succinct* enough to avoid overwhelming the explainee with unnecessary information, i.e., fill in the most gaps with the fewest arguments [26, 33–35, 54]. This is somewhat connected to the novelty (**U8**) of an explanation as it can be partially achieved by avoiding premises that an explainee is already familiar with. Furthermore, parsimony can be used as a tool to reduce complexity (**U9**) of an explanation regardless of the explainee's background knowledge. For example, *brevity* of a counterfactual explanation can be achieved by giving as few reasons (number of conditions) in the statement's foil as possible. This requirement is also related to another rule of cooperative communication presented by Grice et al. [13]. The *maxim of quantity* states that one should only communicate as much information as necessary and no more (a partial explanation).

## 2.4 Safety Requirements

Explainability methods tend to reveal partial information about the data set used to train predictive models, these models' internal mechanics or parameters and their prediction boundaries. Therefore, *Explainability Fact Sheets* aim to consider *the effect of explainability on robustness, security and privacy aspects of predictive systems* which they are built on top of as well as the robustness of explanations themselves.

**S1** *Information Leakage.* Every explainability approach should be accompanied by a critical evaluation of its privacy and security implications and a discussion about mitigating these factors. It is important to consider how much information an explanation reveals about the underlying model and its training data. For example, consider a counterfactual explanation applied to a logical machine learning model; given that this model family applies precise thresholds to data features, this type of an explanation is likely to leak them. On the other hand, explanations for a $k$-nearest neighbours model can reveal training data points and for a support vector machine these could be data points on the support vectors. We could partially mitigate this threat by increasing the parsimony of explanations (**U11**), producing explanations for aggregated data or obfuscating the exact thresholds or data points, for example, by $k$-anonymising [47] the data, outputting fuzzy thresholds in the explanations or providing general directions of change (e.g., "slightly more") to avoid giving out the exact values. Another example can be a security trade-off between *ante-hoc* approaches that reveal information about the predictive model itself and local *post-hoc* explanations that can only leak behaviour of the decision boundary in the neighbourhood of a selected point.

**S2** *Explanation Misuse.* With information leakage in mind, one can ask how many explanations and of how many different data points does it take to gather enough insight to steal or game the underlying predictive model. This can be a major concern especially if the predictive model is a trade secret. Furthermore, explanations can be used by adversaries to game a model; consider a case where a malicious user was able to find a bug in the model by inspecting its explanations, hence is now able to take advantage of it. This observation indicates a close relation between explanations and adversarial attacks. This requirement is closely linked to a consideration of the intended application of an explainability system (**O7**): a system designed as a certification tool will usually reveal more information than one providing explanations to customers. Therefore, having clear target audience in mind (**O6**) is crucial.

**S3** *Explanation Invariance.* Given a phenomenon modelled by a predictive system, data that we gather are just a way to quantify its observed effects. The objective of a predictive system should be to elicit insights about the underlying phenomenon and the explanations are a medium to foster their understanding in a human comprehensible context. Ideally, explanations should be based on a property of the underlying phenomenon rather than an artefact dependent on a predictive model. In this setting it is natural to expect an explainability system to be [15]:

- **consistent** Explanations of similar data points should be similar for a fixed model (training data and procedure) and explanations of a fixed data point should be comparable across different predictive models or different training runs of the same model (trained using the same data).
- **stable** An explainability approach should provide the same explanation given the same inputs (model and/or data point). This could be measured by investigating variance of an explanation over multiple executions of an explainability algorithm.

Ideally, explanations produced by one method should be comparable to these produced using another explainability technique (given fixed training data). If one of these properties is not the case, the designer of an explainability algorithm should investigate how model configuration and parameterisation influence explanations that it produces. Such inconsistency – where the same event is given different, often contradictory, explanations by different actors (explainable algorithms in our case) – is well documented in social sciences as "The Rashmon Effect" [6] and should be avoided.

**S4** *Explanation Quality.* The final safety requirement concerns evaluating the quality and correctness of an explanation with respect to the "confidence" of the underlying predictive model and the distribution of the (training) data. This requirement is in place as poor predictive performance, whether overall or for specific data points, usually leads to uninformative explanations. After all, if a prediction is of poor quality, it would be unreasonable to expect the explanation to be sensible. We suggest that an explanation be accompanied by a list of uncertainty sources, one of which may be the confidence of the predictive model for the instance being explained [39]. For example, if a method relies on synthetic data (as opposed to real data) this should be clearly stated as a source of variability, hence uncertainty. Another example of an explanation that does not convey its quality can be a counterfactual that lies in a sparse region as compared to the distribution of the data set used to train the model. Since we have not seen many data points in that region, we should not trust the explanation without further investigation [30, 40].

## 2.5 Validation Requirements

Finally, *explainable systems should be validated* with user studies (**V1**) or synthetic experiments (**V2**) *in a setting similar to the intended deployment scenario*. This research area has seen increasing interest in the recent years with Doshi-Velez and Kim [7] providing evaluation criteria and various approaches to validate an explainable system. Other researchers [41, 52] highlighted the importance of considering the stakeholders of an explanation before validating it – intended application (**O7**) and audience (**O6**). Nevertheless, there does not appear to be a consensus regarding a validation protocol, which hinders the progress of explainable ML research by making explainable methods incomparable. A commonly agreed validation protocol could help to eliminate confirmation bias (when two explanations are presented side by side), mitigate selection bias (when a study is carried out via Amazon Mechanical Turk all of the participants are computer-literate) and fight a phenomenon called "The Illusion of Explanatory Depth" [46] (to overcome explanation ignorance). For example, when users are asked to choose the best explainability approach out of all the options presented to them, they should not be forced to choose any single one unless they consider at least one of them to be useful. When validating interpretability approaches one should also be aware of a phenomenon called "Change Blindness" [49]: humans inability to notice all of the changes in a presented medium. This is especially important when the explanatory medium (**O2**) is an image. A good practice in such cases is ensuring that two explanations or the instance to be explained and the explanation are clearly distinguishable by, for example, highlighting the change (**U9**).

Furthermore, a well designed user study could also provide a clear answer to some of the (qualitative) explanation properties listed in the previous sections. For example, it can help to evaluate the effectiveness of an explanation for a particular audience (**O6**), assess the background knowledge necessary to benefit from an explanation (**O5**) or check the level of technical skills required to use it (**O2**) as not all explainees may be comfortable with a particular explanatory medium. Standardised user studies are not a silver bullet, however a protocol (such as randomised controlled trails in medical sciences) should be in place given that they are the most acceptable approach to validate the explainability powers of a new method. Doshi-Velez and Kim [7] identified three types of evaluation approaches:

- **application level** Validating an explainability approach on a *real task* with user studies (**V1**), e.g., comparing the explanations given by doctors (*domain experts*) and an explainability algorithm on X-ray images.
- **human level** Validating an explainability approach on a *simplified task* (that represents the same domain) and a lay audience (to avoid using domain experts whose time is often scarce and expensive) with user studies (**V1**), e.g., an Amazon Mechanical Turk experiment asking the explainees to choose the most appealing explanation given multiple different techniques.
- **function level** Validating an explainability approach on a *proxy task* – synthetic validation (**V2**) – e.g., given an already proven explainability method (such as explaining decision trees by visualising their structure), a proxy can be its measure of complexity (tree depth and width).

A different set of, mostly synthetic (**V2**), validation approaches was proposed by Herman [15]:

- using simulated data with known characteristics to validate correctness of explanations, and

- testing stability and consistency of explanations – see the invariance safety requirement (**S3**) for more details.

The latter validation approach can be either *quantitative* (**V2**), given a well-defined metric, or *qualitative* (**V1**), given user's perceptual evaluation.

Lipton [32] has come up with three distinct approaches to evaluate how well an explanation is understood based on user studies (**V2**):

**simultability** Measuring how well a human can recreate or repeat (simulate) the computational process based on provided explanations of a system (for example, by asking the explainee a series of counterfactual what-if questions).

**algorithmic transparency** Measuring the extent to which a human can fully understand a predictive algorithm: its training procedure, provenance of its parameters and the process governing its predictions.

**decomposability** Quantifying the ability of an explainee to comprehend individual parts (and their functionality) of a predictive model: understanding features of the input, parameters of the model (e.g., a monotonic relationship of one of the features) and model's output.

## 3 EXPLAINABLE SYSTEMS TRADE-OFFS

Multiple questions arise when developing an explainability approach for a predictive system and evaluating it based on our list of properties. Are they all of equal importance? Are they compatible with each other or are some of them at odds? In practice, many of them cannot be achieved at the same time and their importance often depends on the *application area* [9]. Moreover, while selected classes of explainability methods (e.g., counterfactuals) may be flexible enough to comply with most of the properties in theory, some of them can be lost in implementation due to particular algorithmic choices. While both *functional* and *operational* requirements are properties of a specific explanation methodology and its implementation, the *usability* desiderata are general properties and any approach should aim to satisfy all of them. For example, making an explainability technique *model-agnostic* forces it to be a *post-hoc* (or a *mimic*) approach and prevents it from taking advantage of specifics of a particular model implementation. Furthermore, such approaches create an extra layer of complexity on top of the predictive model (as opposed to *ante-hoc* techniques) that can be detrimental to *fidelity* of the explanation: a trade-off between *completeness* and *soundness* that is common to *model-agnostic* approaches. Lombrozo [33] points out that explanations that are *simpler* (**U11**), i.e., with fewer causes, *more general* (**U2**) and *coherent* (**U7**) are usually more appealing to humans, however depending on the application (**O7**) and the target audience (**O6**) this may not always be desirable. When considering *coherence* of an explanation we may run into difficulties defining the complement of the concept being explained as *none-concepts* are usually ill-defined, e.g., "What is not-a-car?" Kulesza et al. [28] show that both *completeness* (**U2**) and *soundness* (**U1**) are important, however if faced with a trade-off, one should choose the first over the latter, which, as pointed out by Eiband et al. [9], is not universal and largely depends on the application domain. Similarly, Walton [54] argues that users prefer explanations that are *more plausible* (**U1**), *consistent with multiple outcomes*, i.e., explain many things at once (**U2**), and *simple* (**U9**, **U11**). While daunting, all of these incompatibilities are important to consider as they can help to identify and make informed choices about the trade-offs that every explainability method is facing.

In particular, vanilla counterfactual explanations prioritise *completeness* over *soundness* as they are always data point-specific. Nevertheless, Miller [34] shows that, in theory, counterfactuals – which he considers as the most "human-friendly" explanations since they are contrastive and answer a "Why?" question – can satisfy most of the desiderata and the aforementioned example is an artefact of algorithmic implementations. Moreover, he shows that based on social sciences research some of the properties of explainable systems should be prioritised: *necessary causes* (**U2**, **U3**) are preferred to *sufficient* ones; *intentional actions* (**U10**, **U7**) form more appealing explanations than those taken without deliberation; the *fact* and the *foil* of a (counterfactual) explanation should be clearly *distinguishable* (**U9**); *short* and *selective* (**U11**) explanations are preferred to *complete* ones; the *social context* (**U10**, **U9**, **O6**) should drive the content and the nature of an explanation; and one explanation covering multiple phenomena (**U2**) is preferred to multiple unique explanations.

Some of these properties can be employed to achieve more than one goal. For example, *completeness* can be partially achieved by having *contextfull* explanations. If an explainability system is not inherently *interactive*, this requirement can be achieved by deploying it within an interactive platform such as a dialogue system for explanations delivered in natural language or interactive web page for visualisations. *Actionability* and *chronology* are usually data set-specific and can be achieved by manually annotating features that are actionable, and ordering the time-sensitive ones. *Personalisation* – along with *coherence*, *novelty* and *complexity* that all depend on it – is the most difficult criterion to be satisfied. On one hand, we can argue that the *complexity* (as well as *novelty* and *coherence*) of an explanation can be adjusted by personalising it either by system design (**O2**, **O6**, **O7**), through user interaction (**O3**, **U4**) or with parsimony (**U11**). Alternatively, we can imagine encoding a hierarchy of explanation complexity (based on the user's mental model) and utilising this heuristic to serve explanations of desired complexity.

## 4 DISCUSSION

Systematically evaluating properties of explainability techniques can be a useful precursor to user studies (e.g., helping to design them) to show their capabilities and compliance with the best practices in the field. Furthermore, despite theoretical guarantees of selected desiderata for some explainability approaches, these guarantees can be lost in implementation. For example, model-agnostic explainers can render some desiderata difficult to achieve since these approaches cannot take advantage of model-specific aspects of predictive algorithms. LIME [44] has recently been subjected to studies aiming to validate its usability [31, 50, 57], which discovered lack of stability for its explanations (**S3**) and shortcomings of their locality (**U1**), hence raising questions about the validation methods (**V2**) used to evaluate this technique in the first place. We concur that had a list of requirements such as one presented in this paper been available, some of these issues could have been avoided. To support this claim we show an example of such a *Fact Sheet* as a supplementary material distributed with this paper, which closely inspects properties of the LIME algorithm along all the five dimensions.

## 4.1 Target Audience

We argue that a comprehensive list of requirements for explainability systems spanning both computational and social aspects of thereof is needed amid lack of a general consensus in this space among researchers and practitioners. Some papers discuss a subset of the requirements presented here with many of them being scattered throughout explainability literature, hence rendering it difficult to get a coherent and complete view on the matter. Having all of the requirements in one place will benefit *designers* (researchers) and *users* of explainability systems by enabling them to use the *Fact Sheets* to:

- guide their development, implementation, evaluation and comparison in a systematic and consistent manner,
- identify the gaps between their (theoretical) capabilities and divergence from those for a given implementation, and
- quickly grasp properties of a given method to choose an appropriate approach for a desired use case given the unified design of the *Fact Sheets* (as with food nutrition labels [56] the users know what to expect and how to navigate them).

In addition to serving these two communities and associated with them use cases, our *Fact Sheets* can be utilised as a reporting medium aimed at regulators and certification bodies by providing them with the necessary information in a structured and comparable way. Given the structure of our *Fact Sheets*, browsing through them should be sufficient to make explainability systems more appealing and "transparent" to a wider public, e.g., a non-technical audience.

## 4.2 Delivery Format and Medium

We chose to present the explainability requirements in the form of a (self-reported) fact sheet because we want to empower the designers and users of explainability algorithms to make better choices and consider a wide spectrum of functional, operational, usability, safety and validation aspects when building or using them. Such a flexible and comprehensive structure makes them suitable for a wide range of applications and allows their users to take as much, or as little, as they want from them rather than feel obliged to complete them in their full length and complexity. To indicate this flexibility we opted for the use of the "fact sheet" term as opposed to using "standards", "guidelines" or "recommendations". We suggest that our requirements list can form a future basis of such standards or recommendations, however we are not in a position to enforce this, hence we leave this task to bodies such as IEEE and their Global Initiative on Ethics of Autonomous and Intelligent Systems, which has already produced recommendations for Transparency and Autonomy [37]. Furthermore, posing our requirements list as a "standard reporting tool" could undermine its adoption as enforcing standards may impede the pace of explainability research.

In addition to clear, transparent and well defined structure and content, framing our requirements list as a "fact sheet" has one major advantage: it can be used as a development and deployment *checklist* for explainability approaches. It has been shown that even experienced practitioners can make obvious mistakes, despite their presence of mind, especially under stressful conditions or simply due to the repetitive nature of a task. Checklists have been shown to eradicate most of such trivial, and often dangerous, human errors, for example, a checklist that helps to account for all the tools used during a surgical procedure after it is finished [55]. A similar line of reasoning can be applied to designing and deploying explainability algorithms, however instead of a checklist the user is provided with a "fact sheet" to aid critical evaluation of explainability algorithm capabilities and draw attention to its features that may have been overlooked.

Given the evolving nature of our *Fact Sheet* requirements list and the *Fact Sheets* themselves, we propose to host them on-line within a single repository, e.g., a web page hosted on GitHub. Since the requirements can change over time as new research is published, hosting the *Fact Sheet* template and explainability method-specific *Fact Sheets* on-line will enable their natural evolution, versioning, dissemination and revision supported by community contributions (which can be peer-reviewed following a process similar to OpenReview). All in all, we hope that such collection of *Fact Sheets* and guidelines for their creation will become a go-to resource for learning about transparency of predictive systems.

## 4.3 Creation Responsibility

The effort required to create such a *Fact Sheet* may seem prohibitively time consuming, hence hindering their widespread adoption, however their creation can be incremental and is not limited to the creator of an explainability technique. Moreover, creation of these *Fact Sheets* is not required during the development of explainability techniques since they can be composed post-hoc. Nevertheless, we suggest using our *Fact Sheet* template throughout the development of such approaches as a guideline and a *checklist* to ensure best practices. The process of their creation can be further sped up by allowing the entire explainable AI community to contribute, which may even lead to improving the method itself by spotting its straightforward extensions. All of this is possible because our *Fact Sheets* can be retrofitted since they only require familiarity with the approach or its implementation unlike similar solutions for data sets (requiring knowledge of the data collection process) or AI services (usually treated as trade secrets). All things considered, we argue that researchers designing such methods and software engineers implementing them are best suited (and would benefit most) from composing *Explainability Fact Sheets*.

## 4.4 Dimensions and Requirements Selection

Some of the requirements may seem very similar or strongly related to one another at first glance, hence their choice may seem arbitrary with some arguing to merge or reorganise them. One reason for repetitiveness of selected concepts is the fact that the requirements span five different dimensions with some of them being presented from a social sciences perspective (e.g., users' perception) while others are rather technical (e.g., deployment, performance). Another reason for the fine detail is ensuring versatility of our *Fact Sheets*: in their current form they are applicable to both *inherently transparent predictive algorithms* as well as *standalone interpretability techniques*. One could imagine a *Fact Sheet* discussing how to interpret linear models with their parameters, thereby highlighting caveats such as feature normalisation and independence assumptions. Such an elaborate structure will also allow the users to quickly browse through the headings without delving into their details, hence make them more appealing and accessible.

Many requirements presented in this paper originate from a diverse XAI literature (academic and on-line articles) and have proven to be of value in multiple instances (both theoretical and practical). The higher-level categorisation (dimensions) is role-driven, e.g., deployment, certification and users' perception. When composing the list we were as comprehensive as possible (to avoid bias) and our intervention was limited to grouping together similar concepts presented under different names. All things considered, we acknowledge that our requirements list by no means should be treated as final and definitive. We hope to validate and revise it over time based on the feedback provided by its users (who create and deploy explainability solutions) and the XAI community.

Despite a well defined list of requirements, preparing an exhaustive *Fact Sheet* for any single explainability approach is a labour- and time-consuming challenge. While some of these properties are purely *analytical*, others are *empirical*. We hence identify two approaches to validate them: *quantitative* – that should be measured or can be precisely and definitely answered by assertion, and *qualitative* – that should be defined in the context of a given explainability approach and either justified by a critical discussion (informal argument) or validated with user studies given that they may lack a unique answer, be subjective and be difficult to measure. This lack of a "correct" answer, however, should not be held against the *Fact Sheets* as even a qualitative discussion (of fuzzy properties that cannot be operationalised) is of paramount importance in advancing transparency of explainability approaches, hence clarifying their aspects which cannot be directly measured (e.g., describing and justifying the chosen validation procedure allows the users to assess suitability of a given approach).

## 5 RELATED WORK

The recent surge in interpretability and explainability research in AI may suggest that this is a new research topic, but in fact (human) explainability has been an active research area for much longer in the humanities [34]. This observation has encouraged Miller [34] to review interpretable approaches in the AI literature using insights from the social sciences, which show how human-centred design can benefit AI interpretability [35]. To date, a scattered landscape of *properties* desired of explainable systems has been presented in a wide range of literature [14, 17, 20, 25, 26, 28, 32, 48, 52] that proposes to evaluate explanations by defining their properties and interaction protocol. Despite all of these researchers converging towards a coherent list of properties desired of explainable systems, none of them collected and organised such a systematic list to serve as a guideline for other people interested in the topic. At best, some studies [55] discuss a subset of explainable system requirements supported by illustrative examples, however their main aim is to familiarise the readers with the concepts and not provide them with an evaluation framework. Alternatively, organisations such as IEEE attempt to develop standards for transparency of autonomous systems [37]. Other studies select a subset of properties and use them to *evaluate* a selected approach for a given task, for example, Kulesza et al. [28] evaluate interactive visualisations for music recommendations with respect to explanation *fidelity* (*soundness* and *completeness*), and Kulesza et al. [26] examine interactive visualisations for Naïve Bayes classification of emails with respect to *fidelity*, *interaction*, *parsimony* and *actionability*. Lakkaraju et al. [29] mathematically define some of the desiderata, e.g., *soundness* and *completeness*, to enable quantitative evaluation of their interpretability method.

User studies are often considered the gold standard in explainability research. Doshi-Velez and Kim [7] have come up with guidelines and best practices for evaluating effectiveness of explainability techniques from the perspective of user studies and synthetic validation, and others [37, 41, 52] considered validation of explainable approaches by focusing on their audience. Nevertheless, some research suggests that user studies cannot fully assess the effectiveness of an explanation due to a phenomenon called "The Illusion of Explanatory Depth" [46] or they can yield unexpectedly positive results as simply presenting an explanation to people makes them believe that it is more likely to be true than not [22].

Another approach towards clarifying explainability properties in ML is self-reporting and certification. Approaches such as "data statements" [2], "data sheets for data sets" [11] and "nutrition labels for data sets" [16] can help to characterise a data set in a coherent way. Kelley et al. [19] argued for a similar concept ("nutrition labels for privacy") to assess privacy of systems that handle personal (and sensitive) information. All of these methods revolve around recording details about the data themselves, e.g., the units of features, the data collection process and their intended purpose. Other researchers argued for a similar approach for predictive models: "model cards for model reporting" [36], "nutrition labels for rankings" [56] and "algorithmic impact assessment" forms [42]. Finally, Arnold et al. [1] suggested "fact sheets" for ML services to communicate their capabilities, constrains, biases and transparency.

## 6 CONCLUSIONS AND FUTURE WORK

In this paper we collated and discussed a list of *functional*, *operational* and *usability* characteristics of explainability techniques for predictive systems. We also characterised *safety* (security, privacy and robustness) properties of explanations and discussed their *validation* methods, e.g., user studies. We showed how each property can be used to systematically evaluate explainable ML approaches and discussed their trade-offs. Based on these characteristics, we propose that explainability approaches are accompanied and assessed by means of *Explainability Fact Sheets*, an example of which is provided as a supplementary material distributed with this paper.

We are currently working on creating these *Fact Sheets* for popular explainability techniques such as surrogates [5, 43] and counterfactuals [34]. We will provide a review of their implementations to investigate discrepancies between their theoretical and algorithmic capabilities starting with LIME [44] for surrogates (see the supplementary material) and followed by an optimisation method proposed by Wachter et al. [53] for counterfactuals. We will publish them on a website where XAI developers and users can also submit their own *Explainability Fact Sheets*. More broadly, our future work will aim at evaluating explainability trade-offs in more depth.

# Local Interpretable Model-agnostic Explanations

This is an *Explainability Fact Sheet* for Local Interpretable Model-agnostic Explanations (LIME). It is distributed as a supplementary material of the "Explainability Fact Sheets: A Framework for Systematic Assessment of Explainable Approaches" paper (Kacper Sokol and Peter Flach, 2020) published in Conference on Fairness, Accountability, and Transparency (FAT* 2020).

## Approach Characteristic

### Description

Local Interpretable Model-agnostic Explanations (LIME) is a surrogate explainability method that aims to approximate a local decision boundary with a sparse linear model to interpret individual predictions. It was introduced by this paper and the implementation provided by its authors is capable of explaining *tabular*, *image* and *text* data.

### Implementations

| Python |
| --- |
| LIME |
| bLIMEy |

### Citation

```
@inproceedings{lime,
  author    = {Marco Tulio Ribeiro and Sameer Singh and Carlos Guestrin},
  title     = {"Why Should {I} Trust You?": Explaining the Predictions of
               Any Classifier},
  booktitle = {Proceedings of the 22nd {ACM} {SIGKDD} International
               Conference on Knowledge Discovery and Data Mining,
               San Francisco, CA, USA, August 13-17, 2016},
  pages     = {1135--1144},
  year      = {2016},
}
```

### Variants

#### bLIMEy

build LIME yourself -- a modular framework for building custom local surrogate explainers.

```
@article{blimey,
  title={b{LIME}y: {S}urrogate {P}rediction {E}xplanations {B}eyond {LIME}},
  author={Sokol, Kacper and Hepburn, Alexander and Santos-Rodriguez, Raul
          and Flach, Peter},
  journal={2019 Workshop on Human-Centric Machine Learning (HCML 2019) at the
           33rd Conference on Neural Information Processing Systems
           (NeurIPS 2019), Vancouver, Canada},
  note={arXiv preprint arXiv:1910.13016},
  url={https://arxiv.org/abs/1910.13016},
  year={2019}
}
```

# Related Approaches

N/A

# Functional Requirements

## F1: Problem Supervision Level

LIME works with:

- **supervised** predictive algorithms, and
- **semi-supervised** predictive algorithms.

## F2: Problem Type

LIME is designed for:

- **probabilistic classifiers** and supports: *binary* and *multi-class* classification tasks, and
- **regression** problems.

## F3: Explanation Target

LIME can only explain **predictions** of a Machine Learning model.

## F4: Explanation Breadth/Scope

Explanations produced by LIME are **local**.

## F5: Computational Complexity

For every explained data point the LIME algorithm needs to perform the following *computationally* and/or *time* expensive steps with the cost of each one depending on the actual algorithmic component used:

- **Generating an interpretable data representation** may be necessary for some applications. *Tabular data* may be binned to form human-comprehensible features such as "15 < age < 18". *Images* need to be pre-processed to identify super-pixels. *Text* has to be (possibly pre-processed and) transformed into a bag of words representation.
- In order to train a local model to approximate the local behaviour of a global model, we need to **sample data** around the data point being explained. For *tabular data* the sampling algorithm needs to sample data points with the same number of features as in the original data set or if an interpretable representation is used, the same number of features but each one being multinomial with different values indicating different bins defined on that feature. For *images* and *text* the sampling is based on a binary vector of length equal to the number of unique words for text and the number of super-pixels for images.
- Each sampled data point has to be **predicted** with the global model.
- To enforce the locality of an explanation, sampled data are weighted based on their **distance** to the data point being explained, which has to be computed for every generated data point. While for *text* and *images* this is a distance computed on a binary vector; for *tabular data* without interpretable data representation this would most likely be a more computationally-heavy distance calculation procedure, e.g., Euclidean distance.
- A **feature selection** algorithm may be run on *tabular data* to introduce sparsity into the explanation.
- For every data point being explained, a local model has to be **trained** for each explained class as the local model's task is to predict one class vs. the rest.

## F6: Applicable Model Class

The LIME algorithm is **model agnostic**, therefore it works with any predictive model.

The official LIME implementation uses linear **regression** as a local model, therefore for classification tasks the black-box being explained has to be a **probabilistic** model (i.e., output probabilities).

## F7: Relation to the Predictive System

This approach is **post-hoc**, therefore it can be retrofitted to any predictive system.

## F8: Compatible Feature Types

### Tabular Data

Works with both **categorical** and **numerical** features. If an interpretable data representation is used (default behaviour in the implementation), all of the features become categorical (bins) for the purpose of explanation legibility.

### Images

Images are always transformed into an interpretable data representation, namely super-pixels represented as a binary "on/off" vector.

### Text

Text data are always transformed into an interpretable data representation, namely a bag of words represented as a binary "on/off" vector.

## F9: Caveats and Assumptions

By default the LIME implementation discretises tabular data before the sampling procedure, which leads to the sampled data resembling more of a global rather than a local neighbourhood. This is counterbalanced with the data point weighting step based on the proximity of each data point to the one being explained. Moreover, discretising first means that in order to get global model predictions of the sampled data, we need to "un-discretise" them, which in the LIME implementation is performed by uniformly sampling data from each bin, therefore introducing another source of randomness.

For more details please see "bLIMEy: Surrogate Prediction Explanations Beyond LIME" by Kacper Sokol, et al.

# Operational Requirements

## O1: Explanation Family

**Associations between antecedent and consequent.**

### Tabular Data

The explanations produced by tabular LIME are **associations between antecedent and consequent** -- each feature, or a particular bin on that feature if data are transformed into an interpretable representation, is assigned a positive or a negative influence on the local prediction of a selected class.

### Images and Text

The explanations produced by image and text LIME are **associations between antecedent and consequent** -- each word or super-pixel is assigned a positive or a negative influence on the local prediction of a selected class.

## O2: Explanatory Medium

The explanations are delivered as **visualisations**. For *tabular data* this is feature importance, e.g.:

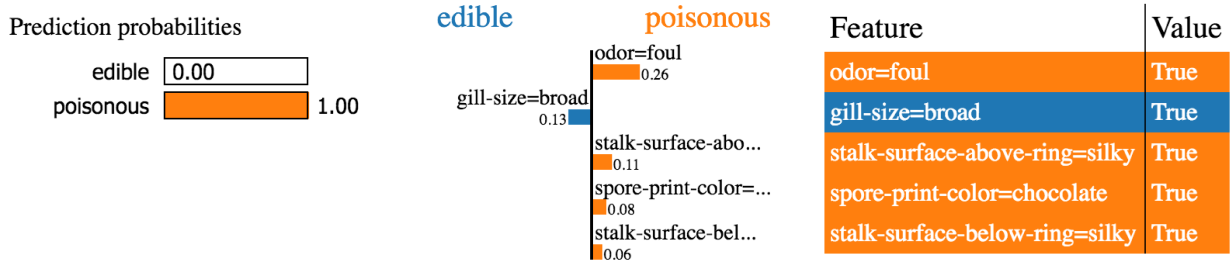

For *text* this is word importance, e.g.:

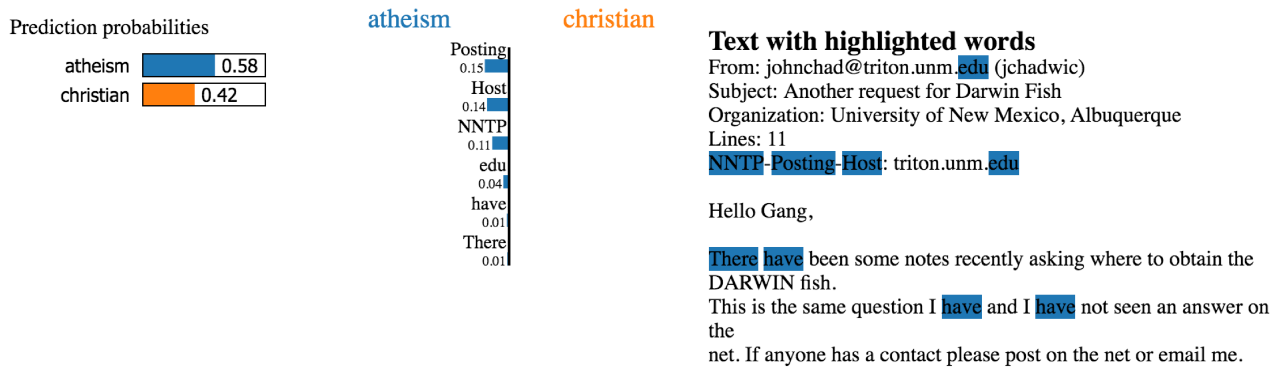

And, for *images* this is super-pixel importance, e.g.:

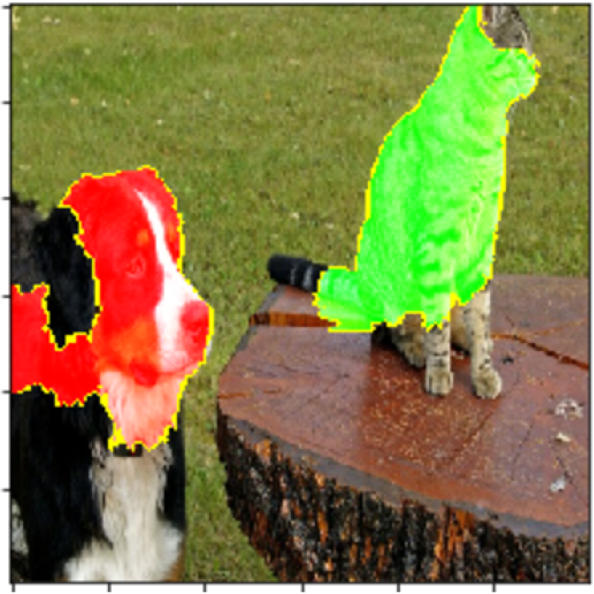

(The figures were taken from the LIME documentation.)

### O3: System Interaction

The explanations are **static** visualisations.

### O4: Explanation Domain

The explanations are expressed as *local* feature importance, i.e., parameters extracted from a locally fitted linear model. For tabular data the explanation can either be in the original domain (training data features) or in an interpretable domain (feature binning). For images these importance factors are expressed for super-pixels and for text these are unique words for a given sentence.

### O5: Data and Model Transparency

For *image* and *text* data there is *no need* for transparency as an interpretable data representation is used. For *tabular data*, regardless of whether an interpretable data representation is used or not, the features *need to* be human-interpretable.

Since this is a *model agnostic* interpretability approach, there is *no need* for the global model to be inherently transparent in any way.

### O6: Explanation Audience

For *tabular data* the audience should be familiar with the general domain of the problem to be able to interpret the meaning of the features. For *images* and *text* any audience is suitable.

The audience is not required to be familiar with Machine Learning concepts.

### O7: Function of the Explanation

The main function of LIME is to increase transparency of a prediction. However, with enough background knowledge the algorithm can also be used as a diagnostic tool when debugging a black-box predictive system.

### O8: Causality vs. Actionability

LIME explanations are **not** of a causal nature. The explanations also lack a direct actionable interpretation.

## O9: Trust vs. Performance

There is **no** performance penalty since LIME is post-hoc and model agnostic. Trust in LIME explanations may suffer given instability and randomness of components influencing the explanation generation process (see **S3** for more details).

## O10: Provenance

LIME explanations are based on *interactions* with the global model and *sampled data*, which affect building a local interpretable linear model. Then, the *parameters* of the local linear model are used as an explanation.

# Usability Requirements

## U1: Soundness

There are two types of local soundness and one type of global soundness that should be measured to evaluate the quality of a LIME explanation. First of all, *mean squared error* (or any other performance metric for numerical values) between a global and a local (used to generate explanations) models should be evaluated in the **neighbourhood** of the instance being explained to understand soundness of the surrogate model around that instance. Then, *mean squared error* in the neighbourhood of the **closest global decision boundary** should be evaluated to understand how well the local model approximates the global decision boundary in that region. Finally, *mean squared error* on the whole data set (e.g., the training data set) should be evaluated to understand overall soundness of the local model.

## U2: Completeness

LIME explanations are **not** complete in their nature. For *images* the explanations are image-specific and for *text* the explanations are sentence-specific. For *tabular data* feature importance cannot be generalised beyond the single data point for which it was generated.

## U3: Contextfullness

Not applicable. LIME explanations do not generalise beyond a data point for which they were generated.

## U4: Interactiveness

The explanations are **static** visualisation. Interactiveness can only be achieved (by technical users) by modifying the interpretable data representation, for example, by adjusting super-pixel boundaries for images.

## U5: Actionability

LIME explanations can only provide importance of a given factor on the black-box decision for a selected data point. They cannot, however, quantify its effect, which the explainee could use to precisely guide his or her future actions.

## U6: Chronology

Chronology is **not** taken into account by LIME explanations.

## U7: Coherence

Coherence is **not** modelled by LIME explanations.

## U8: Novelty

Novelty is **not** considered by LIME explanations.

## U9: Complexity

Complexity of LIME explanations cannot be directly adjusted. It can only be fine-tuned via changes to the interpretable data representation.

## U10: Personalisation

LIME explanations **cannot** be personalised.

## U11: Parsimony

Parsimony is introduced by the **feature selection** step for *tabular data*. Sparsity of *text* and *image* explanations is not necessary as these explanations are overlaid on top of the original image or sentence. For *text* parsimony can be also achieved by presenting the top N words in favour and against a given classification result.

# Safety Requirements

## S1: Information Leakage

Since LIME explanations are expressed in terms of the local model coefficients they do not leak any information. The only leakage that may occur is when creating an interpretable data representation for *tabular data* as some of the discretisation (binning) techniques may reveal characteristics of the data, e.g., quartile binning.

## S2: Explanation Misuse

LIME explanations can be misused by modifying the explained data point according to the feature importance outputted by a local model. Nevertheless, this is not a straightforward task given that the explanation can be expressed in an interpretable data representation. Moreover, this importance is derived for a single data point with a local model, therefore these insights will most likely not generalise beyond that case. Discovering that the same set of factors is important for multiple individual explanations (data points) may be taken advantage of, however given that each insight is derived from a unique local model this is very unlikely.

## S3: Explanation Invariance

LIME explanations may be **unstable** given that the local models are trained on sampled data. To ensure consistency the sampling procedure needs to be controlled either by fixing a random seed or by using a deterministic sampling algorithm. Ideally, the explanations would be imperceptibly different regardless of the data sample. This may be true in the limit of the number of sampled data points, however there is no consideration on the minimum number of samples required to guarantee the explanation stability.

For *images* this also depends on the stochasticity of the segmenter, which generates super-pixels. For *text* the interpretable data representation is deterministic -- a bag of words.

Another source of explanation instability for *tabular data* is the *un-discretisation* step performed by the LIME implementation on sampled data. As it stands, the LIME implementation first discretises the data (to create an interpretable data representation) and then samples within that discretised representation. This means that in order to get predictions of the global model for each sampled data point, they first have to be un-discretised. (For *images* and *text* this is straightforward as the binary vector representation has 1-to-1 mapping with super-pixels in an image or words in a sentence.) The LIME algorithm does that by sampling each feature value from within the bin boundaries, therefore introducing an extra source of randomness to each explanation. For more details please see "bLIMEy: Surrogate Prediction Explanations Beyond LIME" by Kacper Sokol, et al.

No study with respect to consistency of LIME explanations has been carried out.

## S4: Explanation Quality

The quality of an explanation with respect to the confidence of a prediction given by the underlying black-box model or the distribution of the training data is **not** considered.

# Validation Requirements

## V1: User Studies

LIME has been evaluated on three different **user studies**:

1. choosing which of two classifiers generalises better given an explanation,
2. performing feature engineering to improve a model given insights gathered from explanations, and
3. identifying and describing classifier irregularities based on explanations.

Details of these experiments are available in Section 6 of the [LIME paper](#).

## V2: Synthetic Experiments

LIME has been evaluated on three different **simulated** user experiments:

1. validating faithfulness of explanations with respect to the global model -- the agreement between the top K most important features for a glass-box classifier and the top K features chosen by LIME as an explanation,
2. assessing trust in predictions engendered by the explanations -- identifying redundant features, and
3. identifying a better model based on the explanations.

Details of these experiments are available in Section 5 of the [LIME paper](#).